\documentclass{article}

\usepackage[preprint]{neurips_2026}

\workshoptitle{Linguistic Principles for Foundation Models}

\usepackage[utf8]{inputenc}
\usepackage[T1]{fontenc}
\usepackage{hyperref}
\usepackage{url}
\usepackage{booktabs}
\usepackage{amsfonts}
\usepackage{nicefrac}
\usepackage{microtype}
\usepackage{xcolor}
\usepackage{amsmath}
\usepackage{amssymb}
\usepackage{tikz}
\usetikzlibrary{arrows.meta,positioning}
\usepackage{graphicx}

\newcommand{\Addr}{\mathcal{A}}
\newcommand{\Word}{\mathcal{W}}
\newcommand{\Rel}{R}

\title{Compositional Generalization via Structural Identification in a Category-Theoretic Framework}

\author{%
  Akihiro Maeda \\
  The University of Tokyo\\
  Tokyo, Japan\\
  \texttt{akihiromaeda@g.ecc.u-tokyo.ac.jp} \\
  \And
  Thomas Seiller \\
  CNRS \\
  Paris, France \\
  \texttt{thomas.seiller@cnrs.fr} \\
  \And
  Yohei Oseki \\
  The University of Tokyo \\
  Tokyo, Japan \\
  \texttt{oseki@g.ecc.u-tokyo.ac.jp}
}

\begin{document}
\maketitle

\begin{abstract}
Compositional generalization is usually evaluated through model accuracy. 
We instead ask which structural or lexical identifications make held-out COGS examples admissible from the structures observed in training. 
Sentences are represented as functors from syntactic addresses to lexical tokens, and selective collapses induce Kan extensions that propagate observed associations. 
Across 21 COGS generalization types, admissibility follows distinct identification profiles, while residual failures separate unsupported structural templates. 
These data-side diagnoses characterize what the training corpus licenses under specified identifications, without training a predictive model.
\end{abstract}

\section{Introduction}

Large language models generalize broadly across language tasks, yet remain unreliable under systematic distribution shifts involving novel combinations of familiar components.
Compositional generalization concerns this ability to interpret such novel combinations \citep{lake2018generalization}.
COGS \citep{kim2020cogs} is a representative benchmark, including lexical items in novel grammatical roles, familiar structures in unseen combinations, and recursive structures beyond those observed during training.

Model accuracy shows whether a learner succeeds, but not which distinctions in the training data must be identified to generalize a held-out case.
We apply category theory to formalize this process.
Identifying selected structural or lexical distinctions induces a left Kan extension that propagates observed associations across the identified positions.
We hypothesize that compositional-generalization phenomena differ in the identifications sufficient to make them compatible with the training data, and test this hypothesis on COGS.

We represent a sentence as a functor from a category of syntactic addresses to a word category. 
A syntactic collapse identifies addresses, while a lexical collapse identifies token classes.
These collapses determine the Kan extensions tested below.
The Boolean case treated here is the discrete instance of a more general enriched-nucleus construction related to singular-value decomposition~\citep{bradley-2024,gjst-geometric2026}.

Our contribution is an operational characterization of COGS generalization.
Each generalization type is associated with structural or lexical identifications, while unresolved cases reveal whether the obstruction is structural or lexical.
This separates what the training data licenses from what a learner actually realizes.
The resulting operations provide candidate inductive biases for future language models, following the broader program of deriving learning principles from mathematical structure in compositional language models~\citep{maeda2026}.

\section{Categorical formulation of compositional generalization}

\paragraph{COGS as structured data.}
COGS is a semantic parsing benchmark in which each sentence is paired with a logical form specifying predicates and their grammatical relations \citep{kim2020cogs}.
We use these relations to assign each token a syntactic \emph{address}, given by the sequence of dependency labels from the root to that token.
Let
\begin{equation}
  L = \{\texttt{agent},\,\texttt{theme},\,\texttt{recipient},\,
        \texttt{ccomp},\,\texttt{xcomp},\,\texttt{nmod\_in},\,\texttt{nmod\_on},\,
        \texttt{nmod\_beside}\}
\end{equation}
be the finite set of non-root dependency labels used in our COGS representation,
and let $\Addr=L^{*}$ be the set of finite address sequences such as \texttt{ROOT.theme}.
The unit $\epsilon\in\Addr$ denotes the root and is displayed as \texttt{ROOT}.
We equip $\Addr$ with the prefix order
$a\preceq a'$ if there exists $u\in L^{*}$ such that $a'=au$.
A sentence occupies a finite prefix-closed subset $s\subseteq\Addr$: whenever $a'\in s$ and $a\preceq a'$, we also have $a\in s$.
We call $s$ the \emph{template} of the sentence.
A sentence template records which syntactic positions co-occur in a sentence (Appendix~\ref{app:cogs-address}).

\begin{figure}[t]
\centering
\begin{tikzpicture}[
  every node/.style={font=\small},
  lvl/.style={draw,rounded corners=2pt,inner sep=3pt},
  >=Stealth]
\node[lvl] (r)  at (0,2.0)   {\texttt{ROOT}};
\node[lvl] (ag) at (-1.5,1.0){\texttt{agent}};
\node[lvl] (th) at (0.8,1.0) {\texttt{theme}};
\node[lvl] (nm) at (1.4,0.0) {\texttt{theme.nmod\_on}};
\draw (r) -- (ag); \draw (r) -- (th); \draw (th) -- (nm);
\node at (0,2.7) {\textbf{Address tree in }$\Addr$};
\node[lvl] (e) at (6.6,2.0)  {\texttt{eats}};
\node[lvl] (b) at (5.4,1.0)  {\texttt{boy}};
\node[lvl] (c) at (7.4,1.0)  {\texttt{cake}};
\node[lvl] (t) at (8.0,0.0)  {\texttt{table}};
\draw (e) -- (b); \draw (e) -- (c); \draw (c) -- (t);
\node at (6.6,2.7) {\textbf{Word tree in }$\Word$};
\draw[->,dashed] (r)  to[bend left=15] node[above,font=\scriptsize]{$F$} (e);
\draw[->,dashed] (ag) to[bend left=15] (b);
\draw[->,dashed] (th) to[bend right=15] (c);
\draw[->,dashed] (nm) to[bend right=15] (t);
\end{tikzpicture}
\caption{A COGS sentence is represented as a functor $F_x:\Addr_{x}\to\Word_x$. The prefix relation \texttt{theme}$\preceq$\texttt{theme.nmod\_on} is carried to the dependency \texttt{cake}$\to$\texttt{table}. 
A corpus is a family of such functors; for computation, we reduce it to a relation $\Rel\subseteq\Addr\times V$, where $V$ is the vocabulary. 
The syntactic collapse $\sigma$ acts on the address side and the lexical collapse $\tau$ on the lexical side.}
\label{fig:functor}
\end{figure}
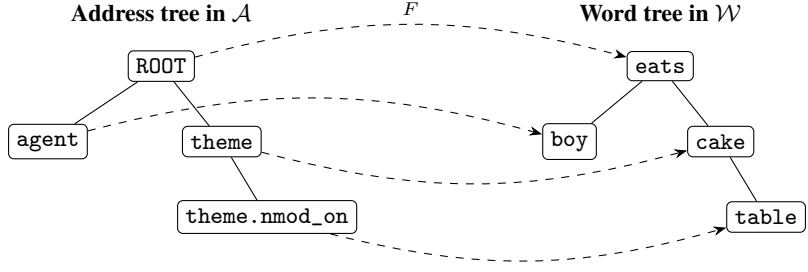

\paragraph{Sentence as a functor.}
For a sentence $x$, let $\mathcal{A}_x$ be the thin category induced by the prefix order on its addresses, and 
let $\mathcal{W}_x$ be the thin category of token occurrences in $x$, ordered by reachability in the corresponding rooted structure.
We represent the sentence as a functor
$
F_x:\mathcal{A}_x \longrightarrow \mathcal{W}_x
\label{eq:sentence-functor}$,
which maps each address to the token occupying that position while preserving the dependency structure (Figure~\ref{fig:functor}). 
The representation separates two aspects of a sentence that are normally observed together: its structural positions and their lexical realizations.

\paragraph{Generalization under controlled identification.}
We model generalization by selectively identifying distinctions in either the structural or lexical representation.
A structural collapse is a map $\sigma:\Addr\to\bar{\Addr}$, where addresses with the same image are treated as equivalent for the purpose of generalization.
Each collapse induces a corresponding functor between the associated thin categories (the categorical construction is given in Appendix~\ref{app:cogs-address}).
The experiments consider several abstractions, including contraction of repeated recursive labels, retention of only the final structural label, and identification of grammatical argument roles.
Similarly, a lexical collapse $\tau : V \to \bar V$ identifies lexical items according to coarser COGS-provided lexical classes.
These operations are not intrinsic linguistic equivalences; they are controlled interventions used to ask which distinctions must be removed for compositional generalization to be realizable (Table~\ref{tab:structural-collapses}). 
\begin{table*}[h]
\caption{Structural identifications used in the analysis. Definitions are given in Appendix~\ref{app:cogs-address}}
\label{tab:structural-collapses}
\centering
\small
\begin{tabular}{lll}
\toprule
Map & Identification & Example \\
\midrule
$\sigma_{\mathrm{id}}$   & No identification & $\texttt{agent}\mapsto\texttt{agent}$ \\
$\sigma_{\mathrm{rec}}$  & Collapse recursive depth & $\texttt{ccomp.ccomp}\mapsto\texttt{ccomp}^{+}$ \\
$\sigma_{\mathrm{last}}$ & Retain only the final relation & $\texttt{ccomp.theme},\texttt{xcomp.theme}\mapsto\texttt{theme}$ \\
$\sigma_{\mathrm{arg}}$  & Identify argument roles & $\texttt{agent},\texttt{theme},\texttt{recipient}\mapsto\texttt{ARG}$ \\
$\sigma_{\mathrm{role}}$ & Identify broad structural roles & $\texttt{nmod\_in},\texttt{nmod\_on},\texttt{nmod\_beside}\mapsto\texttt{NMOD}$ \\
\bottomrule
\end{tabular}
\end{table*}

For computation, we pass from the address category to its underlying set of objects and reduce a training corpus to the incidence relation
$\Rel\subseteq\Addr\times V$, where $V$ is the vocabulary and
$
\Rel(a)=\{w\in V\mid (a,w)\in\Rel\}
\label{eq:corpus-relation}$.
Thus, $\Rel(a)$ records the lexical items observed at syntactic address $a$.
A collapse $\sigma$ groups these assignments by its fibres.
The corresponding left Kan extension gives the fibrewise propagation used to define the admissible lexical set at $a$ as
\begin{equation}
S_a^{+}:=(\operatorname{Lan}_{\sigma}\Rel)(\sigma(a))=\bigcup_{a':\,\sigma(a')=\sigma(a)}\Rel(a').
\label{eq:extended-vocabulary}
\end{equation}
A lexical item observed at one address is therefore locally admissible at another when the selected collapse identifies the two addresses. The collapses $\sigma$ and $\tau$ are externally specified in the present analysis. Thus, the framework characterizes the consequences of a given abstraction rather than inferring the abstraction itself from the corpus. 
For example, under $\sigma_{\mathrm{arg}}$, \texttt{agent} and \texttt{theme} are both mapped to \texttt{ARG}; 
hence, a noun observed only as an agent in training becomes admissible at the theme position after the collapse.
The lexical collapse $\tau$ acts analogously on the lexical axis.
Appendix~\ref{app:kan} gives the categorical formulation.

\paragraph{From local to compositional admissibility.}
The previous construction determines admissible lexical values independently at each structural position.
To test whether a complete held-out sentence is supported, we combine these local admissibility conditions across all positions in its template.
For a sentence template $s$, the admissible complete assignments are
\begin{equation}
\operatorname{Adm}(s)=\prod_{a\in s}S_a^{+}.
\label{eq:admissible-product}
\end{equation}
A held-out sentence is covered only when two conditions are satisfied:
(i) its collapsed structural signature, with occurrence multiplicity retained, is supported by the training data, and
(ii) its lexical assignment belongs to $\operatorname{Adm}(s)$.
These constitute a \emph{structural-template gate} and a \emph{lexical-admissibility gate}, respectively.
The distinction matters empirically: an example can fail because its template remains unsupported even after collapse, or because the template is supported but one or more lexical values remain inadmissible.
Appendix~\ref{app:galois-isbell} gives a closure-theoretic characterization of the Cartesian-product composition in Equation~\ref{eq:admissible-product} and relates its fixed points to formal concepts and the Boolean Isbell nucleus.
It is important that the identifications $\sigma$ and $\tau$ are specified independently of the admissibility computation. The present framework therefore characterizes the consequences of an abstraction rather than learning the abstraction itself. A central direction for future work is to derive such identifications from a compositional type or closure structure.

\section{Empirical Analysis on COGS}

\paragraph{Experimental setup.}
We test this hypothesis across the 21 COGS generalization types.
All address--word associations and structural templates are constructed from the full training split, and coverage is measured deterministically on the generalization split; no predictive model is trained.
We compare the identity map $\sigma_{\mathrm{id}}$, recursion collapse $\sigma_{\mathrm{rec}}$, last-label collapse $\sigma_{\mathrm{last}}$, argument collapse $\sigma_{\mathrm{arg}}$, and $\sigma_{\mathrm{arg}}$ combined with the lexical collapse $\tau$.
A held-out sentence is covered exactly when both the structural-template and lexical-admissibility gates defined in Section~2 are satisfied.

\begin{table*}[t]
\caption{Coverage of COGS generalization types under successive structural and lexical identifications; bold marks the first operation at which each resolved type becomes substantially covered. Per-type results for all 21 cases are reported in Appendix~\ref{app:types}.}
\label{tab:coverage}
\centering
\small
\begin{tabular}{lccccc}
\toprule
Generalization type & id & rec & last & arg & arg $+\tau$ \\
\midrule
Frame alternation (6)
& \textbf{0.878--0.903} & 0.878--0.903 & 0.888--0.914 & 0.890--0.914 & 1.000 \\
Clausal embedding (1)
& 0.000 & 0.010 & \textbf{0.998} & 0.998 & 1.000 \\
Argument-position shift (7)
& 0.000 & 0.000 & 0.000 & \textbf{0.998--1.000} & 1.000 \\
Modifier-position shift (1)
& 0.025 & 0.025 & 0.604 & \textbf{0.920} & 0.922 \\
Primitive-to-role (5)
& 0.000 & 0.000 & 0.000 & 0.000 & \textbf{1.000} \\
Unbounded modifier depth (1)
& 0.000 & 0.003 & 0.220 & 0.220 & 0.222 \\
\midrule
Aggregate (21 types)
& 0.2555 & 0.2561 & 0.3439 & 0.6923 & 0.9592 \\
\bottomrule
\end{tabular}
\end{table*}

\paragraph{Identification profiles.}
As shown in Table~\ref{tab:coverage}, the results support our hypothesis that distinct compositional-generalization phenomena correspond to distinct identification-induced extensions.
Frame alternations are largely admissible without additional identification, whereas clausal embedding becomes almost fully admissible under $\sigma_{\mathrm{last}}$ (0.998).
Argument-position shifts require the stronger $\sigma_{\mathrm{arg}}$ identification (0.998--1.000), while all five primitive-to-role cases remain inadmissible under every tested structural collapse and reach 1.000 only after lexical identification by $\tau$.
Thus, within the tested family of identifications, structural path, argument role, and lexical class provide distinct sufficient abstractions for different generalization phenomena.

\paragraph{Residual structural obstruction.}
Unbounded modifier depth (\texttt{pp\_recursion}) remains largely unresolved under the main identifications, with coverage of only 0.220 under $\sigma_{\mathrm{arg}}$ and 0.222 after adding $\tau$.
The admissibility gates localize this residual failure primarily to unsupported structural templates rather than lexical values.
As shown in Appendix~\ref{app:types}, however, coverage rises to 0.985 under the diagnostic $\sigma_{\mathrm{role}}$, which additionally identifies nominal-modifier relations such as \texttt{nmod\_in}, \texttt{nmod\_on}, and \texttt{nmod\_beside}.
This indicates that generalizing across these modifier relations is required to resolve most of the remaining cases under the tested collapse family.
Such an identification is nevertheless clearly over-coarse: it treats positions introduced by distinct relations such as \emph{in}, \emph{on}, and \emph{beside} as structurally equivalent, and $\sigma_{\mathrm{role}}$ reduces the observed template inventory to only two templates, 
thereby risking overgeneralization to grammatically or semantically inadmissible combinations.
We therefore regard this result as locating the remaining generalization boundary rather than as an appropriate collapse itself.


\section{Prior studies}

The data-side characterization is complementary to the per-type Transformer
accuracies reported by \citet{kim2020cogs}. Coverage measures whether a
held-out example is licensed under a specified identification, whereas
exact-match accuracy measures whether a trained model produces the correct
logical form; the two quantities are therefore not directly comparable.
Their patterns nevertheless differ systematically, with some cases already
licensed by the data despite low model accuracy and others remaining structurally
or lexically blocked (see Appendix~\ref{app:transformer-comparison}).
SLOG and ReCOGS highlight persistent structural-generalization difficulty and sensitivity to logical-form details~\citep{li-etal-2023-slog,wu-etal-2023-recogs}.

Formal concept and lattice structures have also been empirically found in lexical
co-occurrence representations and LLM embedding geometry
\citep{maeda2024,xiong2026}, suggesting a broader connection between
conceptual structure and compositionality.
    


\section{Conclusion and Limitations}

Our analysis supports the hypothesis that, for COGS, compositional generalization can be characterized in terms of the identifications that license unseen combinations through Kan extension.
The resulting operations provide candidate inductive biases for systematic generalization, although we do not train a predictive model here.

A central limitation is that the identifications are externally specified and that the structural and semantic representation is derived from COGS logical forms rather than induced from raw sentences.
One possible direction is to replace the present address--vocabulary relation by a compositional relation on a single space of behaviours, $\Phi\subseteq S\times S$, and construct the nucleus directly from this composition. In such a setting, generated types may provide identifications intrinsically: if $A$ generates a type $E$, then type-theoretic identities  
$A\multimap_* B = E\multimap_* B \hspace{1em} (* \in \{l,r\})$~\citep{gjst-calculus2026} 
express the fact that behaviour on the generators is sufficient to establish behaviour on the generated object. This would turn the present controlled collapses into observable consequences of a compositional type structure, rather than externally specified abstractions. This compositional route is complementary to a second one: replacing the Boolean quantale used throughout by a general value quantale, as in the $\mathbf{R}$-enriched Isbell nuclei of~\citep{gjst-geometric2026}, which would let admissibility be graded rather than binary and would connect syntactic address structure to the projective/tropical geometry of the nucleus.

\bibliographystyle{plainnat}

\appendix

\section{COGS and the address system}
\label{app:cogs-address}

\paragraph{COGS representation.}
COGS \citep{kim2020cogs} is a semantic parsing benchmark in which each source sentence is paired with a logical form specifying lexical predicates and grammatical relations.
For example, a transitive clause contains a verbal predicate together with relations such as \texttt{agent} and \texttt{theme}, while embedded clauses and nominal modifiers introduce relations such as \texttt{ccomp}, \texttt{xcomp}, and \texttt{nmod}.
We use these relations to represent each sentence as a rooted labelled structure and assign each token an address determined by its path from the root.

The COGS dataset and accompanying code are distributed under the MIT License
(Copyright \copyright\ 2020 Najoung Kim).

\paragraph{Address space.}
Let $L$ be the finite set of non-root structural labels used in the COGS representation, including
\texttt{agent}, \texttt{theme}, \texttt{recipient}, \texttt{ccomp}, \texttt{xcomp}, and nominal-modifier labels such as \texttt{nmod\_in} and \texttt{nmod\_on}.
The address space is the free monoid
\begin{equation}
\Addr=L^{*},
\label{eq:address-space}
\end{equation}
whose unit $\epsilon$ denotes the root and is displayed as \texttt{ROOT} in Figure~\ref{fig:functor}.
An address $a=\ell_1\cdots\ell_k$ for $\ell_1,\ldots, \ell_k \in L$ records the sequence of structural labels along the path from the root.
For example, the address of a theme is \texttt{theme}, while a nominal modifier attached to that theme may have address \texttt{theme.nmod\_on}.

We equip $\Addr$ with the prefix order
\begin{equation}
a\preceq a'
\quad\Longleftrightarrow\quad
\exists u\in L^{*}\ \text{such that}\ a'=au.
\label{eq:prefix-order}
\end{equation}
Regarded as a thin category, $\Addr$ contains a unique morphism $a\to a'$ whenever $a\preceq a'$.

\paragraph{Closed sentence templates.}
A sentence template is a finite prefix-closed subset $s\subseteq\Addr$ satisfying
\begin{equation}
a'\in s,\ a\preceq a'
\quad\Longrightarrow\quad
a\in s.
\label{eq:prefix-closed-template}
\end{equation}
Prefix closure guarantees that whenever a structural position occurs, all positions on the path from the root to that position also occur.
Let $\mathcal{S}$ denote the set of such templates and let $\mathcal{S}_{\mathrm{obs}}\subseteq\mathcal{S}$ denote the templates observed in the training corpus.
For a sentence $x$ with template $s_x$, the restriction $\Addr|_{s_x}$ forms its address category, which we denote by $\Addr_x$ by a slight abuse of notation.
Together with its token dependency category $\Word_x$, the sentence is represented by the functor $F_x:\Addr_{x}\to\Word_x$ used in Figure~\ref{fig:functor}.

\paragraph{Corpus relation.}
Let $V$ denote the vocabulary.
A corpus $\{F_x\}_{x\in D}$ induces the incidence relation
\begin{equation}
\Rel=\{(a,w)\in\Addr\times V\mid \text{$w$ is observed at address $a$ in some training sentence}\}.
\label{eq:appendix-corpus-relation}
\end{equation}
Equivalently,
\begin{equation}
\Rel(a)=\{w\in V\mid(a,w)\in\Rel\}.
\label{eq:appendix-address-vocabulary}
\end{equation}
The pair $(\Rel,\mathcal{S}_{\mathrm{obs}})$ is the finite representation used by the generalization procedure.

\paragraph{Structural collapses.}
A structural collapse is a surjective map $\sigma:\Addr\to\bar{\Addr}$ that identifies selected addresses.
For categorical purposes, $\bar{\Addr}$ is equipped with the preorder generated by
$a\preceq a'\Rightarrow\sigma(a)\preceq\sigma(a')$, so that $\sigma$ is a functor.
The experiments use the following collapses.
\begin{itemize}
\item \textbf{Identity, $\sigma_{\mathrm{id}}$.} No structural distinction is removed: $\sigma_{\mathrm{id}}(a)=a$.
\item \textbf{Recursive collapse, $\sigma_{\mathrm{rec}}$.} This collapse removes distinctions in the depth of consecutive repetitions of
the same recursive relation while preserving the remaining address path.
Thus, addresses that differ only in the number of consecutive occurrences of a recursive label are identified.
For example, repeated \texttt{ccomp} depth is represented schematically as
$\texttt{ccomp.ccomp}\mapsto\texttt{ccomp}^{+}$.
The notation $\texttt{ccomp}^{+}$ denotes the resulting recursive-depth class, rather than an additional dependency label.
\item \textbf{Last-label collapse, $\sigma_{\mathrm{last}}$.} Only the final structural label is retained:
\begin{equation}
\sigma_{\mathrm{last}}(\ell_1\cdots\ell_k)=\ell_k\qquad(k\geq1).
\label{eq:sigma-last}
\end{equation}
We set $\sigma_{\mathrm{last}}(\epsilon)=\texttt{ROOT}$. Addresses ending in the same relation are therefore identified regardless of their preceding path.
\item \textbf{Argument collapse, $\sigma_{\mathrm{arg}}$.} Starting from the last-label representation, the grammatical roles \texttt{agent}, \texttt{theme}, and \texttt{recipient} are identified with a common argument class \texttt{ARG}. Thus, positions that differ only in their grammatical argument role become equivalent.
\item \textbf{Role collapse, $\sigma_{\mathrm{role}}$.}
This diagnostic collapse is coarser than $\sigma_{\mathrm{arg}}$ and groups positions into broad structural-role classes.
In particular, nominal-modifier relations such as \texttt{nmod\_in}, \texttt{nmod\_on}, and \texttt{nmod\_beside} are identified with a common modifier class.
Predicate-bearing and argument positions are likewise grouped according to their broad structural roles.
The collapse is used only as an over-coarsening control: it reduces the observed structural inventory to two signatures and is therefore excluded
from the main comparison in Table~\ref{tab:coverage}.
\end{itemize}
These maps need not be interpreted as claims that the corresponding linguistic distinctions are universally equivalent.
They are controlled identification operations used to determine which distinctions must be removed before a held-out COGS combination becomes admissible.
When a collapse is applied to a sentence template, occurrence multiplicity is retained: the collapsed structural signature is the multiset $[\sigma(a)]_{a\in s}$ rather than the ordinary set image $\sigma(s)$. Thus, for example, an \texttt{agent} and a \texttt{theme} may both be labeled \texttt{ARG} while remaining two distinct positions in the sentence.

\paragraph{Lexical collapse.}
The experiment also uses a lexical collapse $\tau:V\to\bar V$.
For each lexical item $w$, the implementation assigns
\begin{equation}
\tau(w)=\bigl(\texttt{node\_kind}(w),\texttt{is\_proper}(w)\bigr).
\label{eq:tau-definition}
\end{equation}
Two lexical items are therefore identified when they share these COGS-provided lexical attributes.
Unlike the structural collapses, $\tau$ should not be interpreted as a learned semantic representation: it is externally supplied category information used to test the effect of lexical identification.

\paragraph{Generalization split.}
The COGS generalization split contains 21 evaluation types.
For reporting, we group them into six frame-alternation types, seven argument-position shifts, one modifier-position shift, one clausal-embedding type, five primitive-to-role types, and one unbounded-modifier-depth type.
Table~\ref{tab:coverage} reports the range of coverage within each group, while the per-type results are given in Appendix~\ref{app:types}.

\section{Kan extensions induced by collapses}
\label{app:kan}

\paragraph{Kan extension.}
We first recall the categorical construction underlying the computation in Section~2.
Let
\[
q:\mathcal{C}\to\mathcal{D},
\qquad
F:\mathcal{C}\to\mathcal{E}
\]
be functors, where $\mathcal{E}$ admits the required colimits.
A \emph{left Kan extension} of $F$ along $q$ is a functor
$\operatorname{Lan}_{q}F:\mathcal{D}\to\mathcal{E}$
together with a natural transformation
\[
\eta:F\Longrightarrow(\operatorname{Lan}_{q}F)\circ q
\]
that is universal among natural transformations
$F\Rightarrow G\circ q$.
When the relevant colimits exist, it is computed pointwise as
\begin{equation}
(\operatorname{Lan}_{q}F)(d)
\cong
\operatorname*{colim}_{(q(c)\to d)\in(q\downarrow d)}
F(c).
\label{eq:general-lan}
\end{equation}
Dually, the right Kan extension is given pointwise by the corresponding limit.

\paragraph{Specialization to the corpus representation.}
The address category in Appendix~\ref{app:cogs-address} carries the prefix order.
For the finite computation in Section~2, however, we retain only its objects:
we pass to the underlying set of addresses and regard that set as a discrete category.
The same convention is applied to the collapsed address set $\bar{\Addr}$.
Section~2 represents the training corpus by the assignment
\begin{equation}
\Rel:\Addr\longrightarrow\mathcal{P}(V),
\qquad
a\longmapsto\Rel(a),
\label{eq:appendix-corpus-assignment}
\end{equation}
where $\mathcal{P}(V)$ is ordered by inclusion and therefore regarded as a
complete thin category.
Thus, at this computational stage, $\Rel$ is a
$\mathcal{P}(V)$-valued functor on the discrete address category.

Precomposition with $\sigma$ gives the reindexing functor
that sends an assignment $\bar U:\bar{\Addr}\to\mathcal{P}(V)$
on the collapsed address set to its pullback along $\sigma$:
\begin{equation}
\Delta_{\sigma}:
\mathcal{P}(V)^{\bar{\Addr}}
\longrightarrow
\mathcal{P}(V)^{\Addr},
\qquad
(\Delta_{\sigma}\bar U)(a)
=
\bar U(\sigma(a)).
\label{eq:delta-reindexing}
\end{equation}
Its left and right adjoints are the left and right Kan extensions along
$\sigma$.  We write
\begin{equation}
\Sigma_{\sigma}
:=
\operatorname{Lan}_{\sigma},
\qquad
\Pi_{\sigma}
:=
\operatorname{Ran}_{\sigma},
\end{equation}
so that
\begin{equation}
\Sigma_{\sigma}
\dashv
\Delta_{\sigma}
\dashv
\Pi_{\sigma}.
\label{eq:sigma-delta-pi}
\end{equation}

\paragraph{Fibre-wise form.}
Because the source and target address categories used in the computation are
discrete, the comma category appearing in
Equation~\ref{eq:general-lan} reduces to the fibre
$\sigma^{-1}(\bar a)$.
Moreover, joins and meets in $\mathcal{P}(V)$ are respectively set union and
intersection.
Hence the two Kan extensions take the explicit form
\begin{equation}
(\Sigma_{\sigma}U)(\bar a)
=
\bigcup_{a\in\sigma^{-1}(\bar a)}U(a),
\qquad
(\Pi_{\sigma}U)(\bar a)
=
\bigcap_{a\in\sigma^{-1}(\bar a)}U(a).
\label{eq:fibre-kan}
\end{equation}
Equivalently, for
$U:\Addr\to\mathcal{P}(V)$ and
$\bar U:\bar{\Addr}\to\mathcal{P}(V)$,
\begin{equation}
\Sigma_{\sigma}U\subseteq\bar U
\iff
U\subseteq\Delta_{\sigma}\bar U,
\qquad
\Delta_{\sigma}\bar U\subseteq U
\iff
\bar U\subseteq\Pi_{\sigma}U,
\label{eq:kan-adjunctions}
\end{equation}
where the inclusions are pointwise.


\paragraph{Extension induced by a structural collapse.}
Applying the unit and counit of these adjunctions to the observed corpus
assignment gives
\begin{equation}
\Delta_{\sigma}\Pi_{\sigma}\Rel
\subseteq
\Rel
\subseteq
\Delta_{\sigma}\Sigma_{\sigma}\Rel.
\label{eq:kan-bracket}
\end{equation}
The lower assignment retains a lexical item at an address only when it occurs
at every address in the same $\sigma$-fibre.
The experiments use the upper assignment
\begin{equation}
\Rel_{\sigma}^{+}
=
\Delta_{\sigma}\Sigma_{\sigma}\Rel,
\label{eq:upper-extension}
\end{equation}
which is the least fibre-constant assignment containing $\Rel$.
In particular, at an original address $a$,
\begin{equation}
S_a^{+}
=
\Rel_{\sigma}^{+}(a)
=
(\Sigma_{\sigma}\Rel)(\sigma(a))
=
\bigcup_{a':\,\sigma(a')=\sigma(a)}
\Rel(a').
\label{eq:local-admissibility-appendix}
\end{equation}
This is exactly the local admissibility operation used in
Equation~\ref{eq:extended-vocabulary} of the main text.
The newly admitted address--word pairs are therefore
\begin{equation}
N_{\sigma}
=
\Rel_{\sigma}^{+}\setminus\Rel,
\end{equation}
which records combinations licensed by the selected structural identification
but not directly observed in training.

\paragraph{Lexical collapse.}
The same construction applies to the lexical axis.
Transpose the incidence relation as
\begin{equation}
\Rel^{\top}:V\longrightarrow\mathcal{P}(\Addr),
\qquad
\Rel^{\top}(w)
=
\{a\in\Addr\mid(a,w)\in\Rel\}.
\label{eq:transpose-relation}
\end{equation}
For a lexical collapse $\tau:V\to\bar V$, its left Kan extension is
\begin{equation}
(\Sigma_{\tau}\Rel^{\top})(\bar w)
=
\bigcup_{w':\,\tau(w')=\bar w}\Rel^{\top}(w').
\label{eq:lexical-kan}
\end{equation}
Thus, after reindexing to the original vocabulary, a word inherits the address
associations of all lexical items in the same $\tau$-fibre.

When structural and lexical collapses are applied together, the extended
incidence relation on the original address and vocabulary sets is
\begin{equation}
\Rel_{\sigma,\tau}^{+}
=
\left\{
(a,w)\in\Addr\times V
\;\middle|\;
\exists(a',w')\in\Rel:
\sigma(a')=\sigma(a),\;
\tau(w')=\tau(w)
\right\}.
\label{eq:joint-extension}
\end{equation}
The experiments using $\sigma_{\mathrm{arg}}+\tau$ evaluate admissibility with
respect to this jointly extended relation.

\paragraph{Relation to the full address category.}
The Kan extensions used in the present computation are taken after passing
from the address categories to their underlying sets, regarded as discrete
categories.
Consequently, Equation~\ref{eq:fibre-kan} depends only on the fibres of
$\sigma$.
The prefix order on $\Addr$ instead determines sentence templates and the
functorial structure of the original sentence representation.
An order-sensitive extension in which the non-identity morphisms of the
address category also participate in the Kan-extension computation is left
for future work. 

\section{Galois closure and the Boolean Isbell interpretation}
\label{app:galois-isbell}

The Galois closure can equivalently be expressed as the Isbell closure associated with the corresponding Boolean profunctor.
Thus the admissible products are fixed points of a Boolean Isbell closure. In the present discrete setting, this observation is an interpretation of the Cartesian admissibility construction rather than an additional computational operation. 

\paragraph{Sections and coordinate constraints.}
Fix a sentence template $s$ and let
\begin{equation}
\operatorname{Sec}(s)=V^{s}
\label{eq:sections}
\end{equation}
be the set of complete lexical assignments on $s$.
For each address $a\in s$ and subset $S\subseteq V$, define a coordinate
constraint $(a,S)$, and let
\begin{equation}
\mathcal{C}_{s}
=
\{(a,S)\mid a\in s,\;S\subseteq V\}.
\label{eq:coordinate-constraints}
\end{equation}
Define a binary relation
\begin{equation}
I_s
\subseteq
\operatorname{Sec}(s)\times\mathcal{C}_{s}
\end{equation}
by
\begin{equation}
t\, I_s\, (a,S)
\quad\Longleftrightarrow\quad
t(a)\in S.
\label{eq:satisfaction-relation}
\end{equation}
Thus, a section is related to a coordinate constraint exactly when its value
at that coordinate satisfies the constraint.

\paragraph{The induced Galois connection.}
For $X\subseteq\operatorname{Sec}(s)$, define
\begin{equation}
X^{\uparrow}
=
\{c\in\mathcal{C}_{s}
\mid
t\,I_s\,c \text{ for every } t\in X\},
\label{eq:galois-up}
\end{equation}
and for $Y\subseteq\mathcal{C}_{s}$ define
\begin{equation}
Y^{\downarrow}
=
\{t\in\operatorname{Sec}(s)
\mid
t\,I_s\,c \text{ for every } c\in Y\}.
\label{eq:galois-down}
\end{equation}
These operators form an antitone Galois connection:
\begin{equation}
X\subseteq Y^{\downarrow}
\quad\Longleftrightarrow\quad
Y\subseteq X^{\uparrow}.
\label{eq:galois-adjunction}
\end{equation}
Consequently,
$X\mapsto X^{\uparrow\downarrow}$ is a closure operator on
$\mathcal{P}(\operatorname{Sec}(s))$.

A formal concept of the context
$(\operatorname{Sec}(s),\mathcal{C}_{s},I_s)$
is a pair $(X,Y)$ satisfying
\begin{equation}
X^{\uparrow}=Y,
\qquad
Y^{\downarrow}=X.
\end{equation}
Equivalently, its extent $X$ is a fixed point of
$X\mapsto X^{\uparrow\downarrow}$.

\paragraph{Closed sets are rectangular products.}
For $X\subseteq\operatorname{Sec}(s)$, let
\begin{equation}
\pi_a(X)
=
\{t(a)\mid t\in X\}
\subseteq V
\end{equation}
denote its projection onto coordinate $a$.
By Equation~\ref{eq:satisfaction-relation},
\begin{equation}
(a,S)\in X^{\uparrow}
\quad\Longleftrightarrow\quad
\pi_a(X)\subseteq S.
\end{equation}
Therefore a section $t$ belongs to $X^{\uparrow\downarrow}$ exactly when
$t(a)\in\pi_a(X)$ for every $a\in s$, and hence
\begin{equation}
X^{\uparrow\downarrow}
=
\prod_{a\in s}\pi_a(X).
\label{eq:rectangular-closure}
\end{equation}
The closure thus adds precisely the assignments obtained by freely
recombining values already present in the coordinate-wise projections of
$X$.

It follows that the nonempty fixed points of this closure are exactly the
rectangular subsets
\begin{equation}
B
=
\prod_{a\in s}S_a,
\label{eq:rectangular-fixed-point}
\end{equation}
where each $S_a\subseteq V$ is nonempty.

\paragraph{Admissible combinations as formal concepts.}
Appendix~\ref{app:kan} constructs, for each address $a$, the locally
admissible lexical set $S_a^{+}\subseteq V$ by Kan extension.
Section~2 then combines these local sets into complete sentence-level
assignments.
The corresponding set of complete admissible assignments is
\begin{equation}
\operatorname{Adm}(s)
=
\prod_{a\in s}S_a^{+}.
\label{eq:appendix-admissible-set}
\end{equation}
Provided that every $S_a^{+}$ is nonempty,
Equation~\ref{eq:rectangular-fixed-point} shows that
$\operatorname{Adm}(s)$ is a fixed point of
$(-)^{\uparrow\downarrow}$.
Hence it is the extent of a formal concept of
$(\operatorname{Sec}(s),\mathcal{C}_{s},I_s)$.

This construction is distinct from applying formal concept analysis directly
to the raw corpus relation
$\Rel\subseteq\Addr\times V$.
The Kan extension of Appendix~\ref{app:kan} first determines the locally
admissible sets $S_a^{+}$; the Galois closure considered here then
characterizes their Cartesian recombination at the sentence level.

\paragraph{Relation to the Boolean Isbell nucleus.}
Let $\mathbf{2}=\{0<1\}$ be the Boolean quantale, with tensor given by
conjunction.
Regard $\operatorname{Sec}(s)$ and $\mathcal{C}_{s}$ as discrete
$\mathbf{2}$-enriched categories.
Using the convention that a $\mathbf{2}$-enriched profunctor
$\Phi:X\nrightarrow Y$ is a
$\mathbf{2}$-functor $X^{\mathrm{op}}\otimes Y\to\mathbf{2}$,
the relation $I_s$ defines the profunctor
\begin{equation}
\Phi_s:
\operatorname{Sec}(s)^{\mathrm{op}}
\otimes
\mathcal{C}_{s}
\longrightarrow
\mathbf{2},
\qquad
\Phi_s(t,c)=1
\quad\Longleftrightarrow\quad
t\,I_s\,c.
\label{eq:boolean-distributor}
\end{equation}

Under the standard identification of $\mathbf{2}$-valued presheaves and
copresheaves on discrete categories with subsets, the Isbell conjugacy induced
by $\Phi_s$ acts as
\begin{equation}
X
\longmapsto
X^{\uparrow},
\qquad
Y
\longmapsto
Y^{\downarrow},
\label{eq:isbell-galois}
\end{equation}
with $(-)^{\uparrow}$ and $(-)^{\downarrow}$ exactly as defined in
Equations~\ref{eq:galois-up} and~\ref{eq:galois-down}.
Thus the closure induced on
$\mathcal{P}(\operatorname{Sec}(s))$ is precisely
\begin{equation}
X
\longmapsto
X^{\uparrow\downarrow}.
\end{equation}

The fixed points of the closure
$X \mapsto X^{\uparrow\downarrow}$
therefore form the Boolean Isbell nucleus associated with $\Phi_s$.
Each such fixed extent $X$ determines the corresponding intent
$X^{\uparrow}$, recovering exactly the formal concepts defined above.
In particular, the admissible products
$\operatorname{Adm}(s)$ used in the experiment are fixed points of the
associated closure and hence objects of this nucleus.

\paragraph{Beyond Boolean admissibility.}
The present construction uses the Boolean quantale $\mathbf{2}$, so
admissibility is binary.
Replacing $\mathbf{2}$ by a suitable quantale would replace Boolean incidence
by graded compatibility while retaining the profunctor--Isbell construction, recovering the $\mathbf{R}$-enriched Isbell nuclei of \citep{gjst-calculus2026,gjst-geometric2026}.

This provides the categorical direction for incorporating graded semantic or
selectional relations discussed in Section~5.

\section{Per-type coverage for all generalization cases}
\label{app:types}

\paragraph{Diagnostic gates.}
Table~\ref{tab:types} reports coverage separately for all 21 COGS generalization types underlying the grouped results in Table~\ref{tab:coverage}.
The implementation distinguishes three diagnostic failure modes: \textbf{G1}, an address remains outside the collapsed training axis; \textbf{G2}, the collapsed structural template is unsupported; and \textbf{G3}, the template is supported but at least one lexical item is outside the admissible vocabulary.
The gate column reports the dominant residual obstruction after $\sigma_{\mathrm{arg}}$ only for types whose coverage remains below 0.5 at that stage; \texttt{--} therefore means that no gate label is assigned in this summary.

\begin{table}[t]
\caption{Per-type coverage for all 21 generalization cases. The \emph{gate} column records the dominant residual obstruction after the argument collapse for types with coverage below 0.5: G1 denotes an unavailable collapsed address, G2 an unsupported structural template, and G3 an unsupported lexical value. All five primitive-to-role cases exhibit G3 failures and are resolved by $\tau$, whereas \texttt{pp\_recursion} exhibits a G2 failure and remains largely unresolved. $\sigma_{\mathrm{role}}$ is shown only as an over-coarsening control.}
\label{tab:types}
\centering
\scriptsize
\setlength{\tabcolsep}{2.2pt}
\begin{tabular}{lrrrrrrc}
\toprule
type & id & rec & last & arg & role & arg+$\tau$ & gate\\
\midrule
passive\_to\_active & .878 & .878 & .888 & .890 & 1.000 & 1.000 & --\\
active\_to\_passive & .895 & .895 & .895 & .895 & 1.000 & 1.000 & --\\
unacc\_to\_transitive & .885 & .885 & .896 & .898 & 1.000 & 1.000 & --\\
obj\_omitted\_transitive\_to\_transitive & .885 & .885 & .896 & .898 & 1.000 & 1.000 & --\\
do\_dative\_to\_pp\_dative & .894 & .894 & .911 & .911 & .999 & 1.000 & --\\
pp\_dative\_to\_do\_dative & .903 & .903 & .914 & .914 & .999 & 1.000 & --\\
cp\_recursion & .000 & .010 & .998 & .998 & 1.000 & 1.000 & --\\
obj\_pp\_to\_subj\_pp & .025 & .025 & .604 & .920 & 1.000 & .922 & --\\
pp\_recursion & .000 & .003 & .220 & .220 & .985 & .222 & G2\\
subj\_to\_obj\_common & .000 & .000 & .000 & .998 & .998 & 1.000 & --\\
subj\_to\_obj\_proper & .000 & .000 & .000 & 1.000 & 1.000 & 1.000 & --\\
obj\_to\_subj\_common & .000 & .000 & .000 & .998 & 1.000 & 1.000 & --\\
obj\_to\_subj\_proper & .000 & .000 & .000 & .998 & .999 & 1.000 & --\\
only\_seen\_as\_transitive\_subj\_as\_unacc\_subj & .000 & .000 & .000 & 1.000 & 1.000 & 1.000 & --\\
only\_seen\_as\_unacc\_subj\_as\_obj\_omitted\_transitive\_subj & .000 & .000 & .000 & 1.000 & 1.000 & 1.000 & --\\
only\_seen\_as\_unacc\_subj\_as\_unerg\_subj & .000 & .000 & .000 & 1.000 & 1.000 & 1.000 & --\\
prim\_to\_subj\_common & .000 & .000 & .000 & .000 & .000 & 1.000 & G3\\
prim\_to\_subj\_proper & .000 & .000 & .000 & .000 & .000 & 1.000 & G3\\
prim\_to\_obj\_common & .000 & .000 & .000 & .000 & .000 & 1.000 & G3\\
prim\_to\_obj\_proper & .000 & .000 & .000 & .000 & .000 & 1.000 & G3\\
prim\_to\_inf\_arg & .000 & .000 & .000 & .000 & .000 & 1.000 & G3\\
\bottomrule
\end{tabular}
\end{table}

\paragraph{Per-type structure of the result.}
The full results confirm that the grouped patterns in Table~\ref{tab:coverage} are not produced by averaging heterogeneous cases.
All seven argument-position shifts move from zero coverage under $\sigma_{\mathrm{last}}$ to approximately complete coverage under $\sigma_{\mathrm{arg}}$.
Likewise, all five primitive-to-role cases remain at zero under every structural collapse and reach complete coverage only after applying $\tau$.
The two recursion cases behave differently: \texttt{cp\_recursion} is almost completely resolved by $\sigma_{\mathrm{last}}$, whereas \texttt{pp\_recursion} remains limited by its unsupported template structure.
The per-type analysis therefore reinforces the central observation that different compositional generalizations are associated with different data-side obstructions and identification operations.

\paragraph{Progressive abstraction and over-coarsening.}
At the aggregate level, coverage increases from 0.2555 under the identity map to 0.3439 under $\sigma_{\mathrm{last}}$, 0.6923 under $\sigma_{\mathrm{arg}}$, and 0.9592 after lexical identification by $\tau$.
This increase is not uniform across generalization types: different groups become admissible at different stages of abstraction.
Coverage alone is therefore not a criterion for selecting an appropriate collapse.
In particular, the diagnostic $\sigma_{\mathrm{role}}$ attains aggregate coverage of 0.7610 while reducing the observed structural inventory to only two templates.
We consequently treat it as an over-coarsening control rather than as a candidate generalization operation.

\subsection{Comparison with the prior study}
\label{app:transformer-comparison}

Table~\ref{tab:transformer-comparison} compares the data-side characterization obtained from the proposed identifications with the grouped Transformer accuracies reported by \citet{kim2020cogs}.
The quantities have different meanings and are not intended for numerical comparison; the table is included to show how the operational profiles identified in our analysis relate to previously reported model behavior.

\begin{table}[h]
\caption{Data-side characterization versus published Transformer accuracy \citep{kim2020cogs}.}
\label{tab:transformer-comparison}
\centering
\small
\begin{tabular}{lcc}
\toprule
Generalization type & Associated identification & Transformer \\
\midrule
Frame alternation (6) & none & 0.38--0.99 \\
Argument-position shift (7) & $\sigma_{\mathrm{arg}}$ & 0.30--0.87 \\
Clausal embedding (1) & $\sigma_{\mathrm{last}}$ & 0.00 \\
Modifier-position shift (1) & $\sigma_{\mathrm{arg}}$ & 0.00 \\
Primitive-to-role (5) & $\tau$ & 0.00--0.17 \\
Unbounded modifier depth (1) & unresolved & 0.00 \\
\bottomrule
\end{tabular}
\end{table}

\section*{Use of LLM-based tools.}
LLM-based assistants were used to support literature exploration, discussion
and refinement of categorical formulations, and checking mathematical
derivations. All mathematical claims, references, experimental procedures,
and conclusions were independently verified and remain the responsibility
of the authors.


\begin{thebibliography}{9}
\small
\bibitem[Bradley et al.(2024)]{bradley-2024}
T.~D.~Bradley, J.~L.~Gastaldi, and J.~Terilla.
The structure of meaning in language: parallel narratives in linear algebra and category theory.
\emph{Notices of the American Mathematical Society}, 71(2):174--185, February 2024.

\bibitem[Gastaldi et al.(2026a)]{gjst-calculus2026}
J.~L.~Gastaldi, S.~Jarvis, T.~Seiller, and J.~Terilla.
A calculus of types in Isbell nuclei.
arXiv:2606.03369, 2026a.

\bibitem[Gastaldi et al.(2026b)]{gjst-geometric2026}
J.~L.~Gastaldi, S.~Jarvis, T.~Seiller, and J.~Terilla.
Projective metric geometry of tropical nuclei: gap matrices, event loci, and order chambers.
arXiv:2601.07900, 2026b.

\bibitem[Kim and Linzen(2020)]{kim2020cogs}
N.~Kim and T.~Linzen.
COGS: A compositional generalization challenge based on semantic interpretation.
\emph{Proceedings of EMNLP}, 2020.

\bibitem[Lake and Baroni(2018)]{lake2018generalization}
B.~M.~Lake and M.~Baroni.
Generalization without systematicity: On the compositional skills of sequence-to-sequence recurrent networks.
\emph{Proceedings of ICML}, 2018.

\bibitem[Li et al.(2023)]{li-etal-2023-slog}
B.~Li, L.~Donatelli, A.~Koller, T.~Linzen, Y.~Yao, and N.~Kim.
SLOG: A structural generalization benchmark for semantic parsing.
In \emph{Proceedings of the 2023 Conference on Empirical Methods in Natural Language Processing}, pages 3213--3232, 2023.


\bibitem[Maeda et~al.(2024)]{maeda2024}
A.~Maeda, T.~Torii, and S.~Hidaka. 
Decomposing co-occurrence matrices into interpretable components as formal concepts.
\emph{Findings of the Association for Computational Linguistics: ACL}, 2024.

\bibitem[Maeda et al.(2026)]{maeda2026}
A.~Maeda, T.~Torii, Y.~Oseki, and S.~Hidaka.
Mathematical foundations of compositional language models: learning as an inverse problem in representation theory for compositional generalization.
\emph{Transactions of the Japanese Society for Artificial Intelligence}, 41(4):AN40--A\_1, 2026.

\bibitem[Wu et al.(2023)]{wu-etal-2023-recogs}
Z.~Wu, C.~D.~Manning, and C.~Potts.
ReCOGS: How incidental details of a logical form overshadow an evaluation of semantic interpretation.
\emph{Transactions of the Association for Computational Linguistics}, 11:1719--1733, 2023.


\bibitem[Xiong(2026)]{xiong2026}
B.~Xiong.
\newblock The lattice representation hypothesis of large language models.
\newblock In \emph{International Conference on Learning Representations},
  2026.

\end{thebibliography}
\end{document}